\documentclass[letterpaper, 10 pt, journal, twoside]{IEEEtran}

\usepackage{graphicx}
\usepackage{subfig}
\usepackage{amsmath}
\let\labelindent\relax
\usepackage{enumitem}
\usepackage{amssymb}
\usepackage{float}
\usepackage[ruled,vlined,linesnumbered]{algorithm2e}

\newcommand{\xvec}{ \bar{ r }}
\newcommand{\thvec}{\bar{ \theta }}

\usepackage{tikz}
\usepackage{textcomp}
\usepackage{hyperref}
\usepackage{lipsum}

\newcommand\copyrighttext{
  \footnotesize \textcopyright © 2021 IEEE.  Personal use of this material is permitted.  Permission from IEEE must be obtained for all other uses, in any current or future media, including reprinting/republishing this material for advertising or promotional purposes, creating new collective works, for resale or redistribution to servers or lists, or reuse of any copyrighted component of this work in other works.}
\newcommand\copyrightnotice{
\begin{tikzpicture}[remember picture,overlay]
\node[anchor=south,yshift=10pt] at (current page.south) {\fbox{\parbox{\dimexpr\textwidth-\fboxsep-\fboxrule\relax}{\copyrighttext}}};
\end{tikzpicture}%
}

\begin{document}
\copyrightnotice
\title{InsertionNet - A Scalable Solution for Insertion}

\author{Oren Spector$^{1}$ and Dotan Di~Castro$^{1}$
\thanks{$^{1}$ The authors are with Bosch Center Of Artificial Intelligence, Haifa, Israel($\textrm{ email:}${\tt\small oren.spector@gmail.com; dotan.dicastro@gmail.com})}
        }

\makeatletter
\g@addto@macro\@maketitle{
  \begin{figure}[H]
  \setlength{\linewidth}{\textwidth}
  \setlength{\hsize}{\textwidth}
  \subfloat[]{\includegraphics[width=0.774\linewidth]{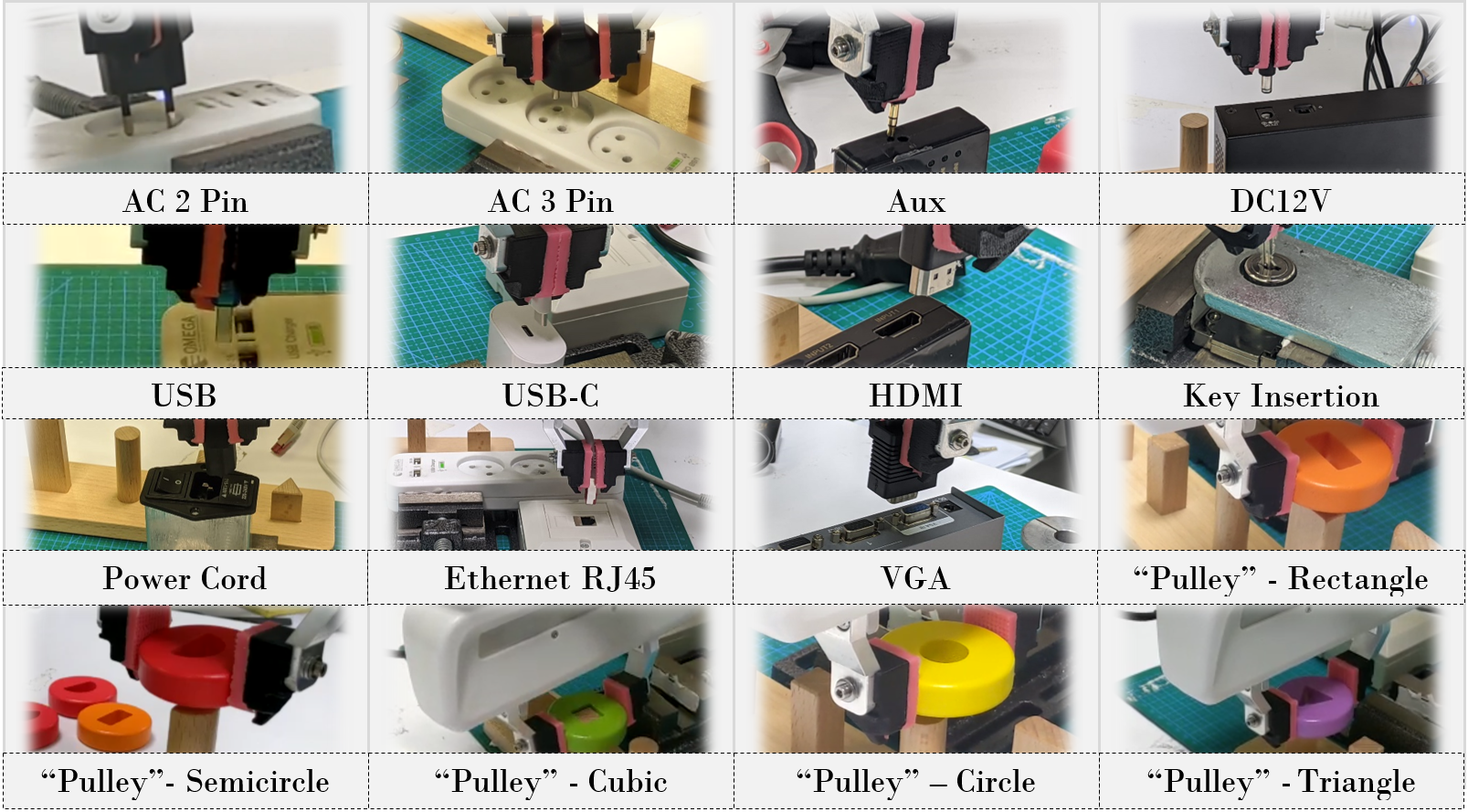}}
     \hfill
      \subfloat[]{\includegraphics[width=0.22\linewidth]{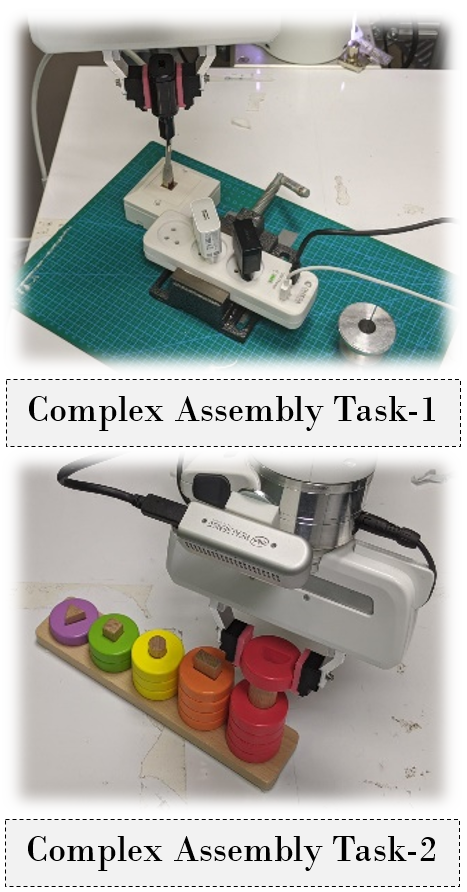}}
     \caption{\label{fig:scalable_solution}(a) 16 different real-life insertion and threading tasks with different properties. (b) complex assembly with consecutive insertion tasks from grasping to holes localization to pegs insertion}
  \end{figure}
}



\maketitle


\begin{abstract}
Complicated  assembly  processes  can  be described as a sequence of two main activities: grasping and insertion. While general grasping solutions are common in industry, insertion is  still only applicable to small subsets of problems, mainly ones involving simple shapes in fixed locations and in which the variations are not taken into consideration. Recently, RL approaches with prior knowledge (e.g., LfD or residual policy) have been adopted. However, these approaches might be problematic in contact-rich tasks since interaction might endanger the robot and its equipment.
In this paper, we tackled this challenge by formulating the problem as a regression problem. By combining visual and force inputs, we demonstrate that our method can scale to 16 different insertion tasks in less than 10 minutes. The resulting policies are robust to changes in the socket position, orientation or peg color, as well as to small differences in peg shape. Finally, we demonstrate an end-to-end solution for 2 complex assembly tasks with multi-insertion objectives when the assembly board is randomly placed on a table. 
\end{abstract}

\section{Introduction}
\label{s:INTRO}

\IEEEPARstart{T}{oday's} assembly tasks still rely heavily on predefined actions or movements and are limited in their ability to adapt to even small changes. This is due to the great variability in the types of problems at hand that differ in many features. Furthermore, state uncertainties in grasp alignment and hole location may result in hazardous forces due to collision with the hole surface. 
Using grasping and insertion together enables us to achieve most assembly tasks.

%

Insertion, despite the large body of work on it, is still only applicable to small subsets of tasks. Most solutions address round objects in a fixed location and do not take industrial variations into consideration. This makes current algorithms, such as the popular spiral search, unable to quickly adapt to new tasks, and they also require specific engineering for each use-case, and are costly.

Recently, researchers have tried to tackle this problem by combining reinforcement learning (RL) algorithms, known for their versatility and generality, with prior knowledge on the system, for example, learning from demonstration (Lfd) or learning only a residual policy . These methods involve learning algorithms such as off policy RL  \cite{vecerik2019practical}, \cite{lee2020guided} and model-based RL \cite{luo2018deep}, which are performed directly on the robot. While achieving interesting results in complex insertion tasks and demonstrating generalization properties to some extent \cite{lee2019making}, this type of online interaction is impractical on contact-rich tasks, either because data collection on real robots is extremely expensive or because unpredicted moves created by initial policies can endanger the robot and its equipment (e.g., force torque sensor), since hazardous contact forces could arise during the trials. Another RL approach is to combine a high-quality simulation with some sim-to-real algorithm \cite{spector2020deep}. While this approach can achieve generalization over different insertion tasks, its scalability is bounded by the ability to model the world via simulation.
In this paper, we formulate the problem differently and solve it as a regression problem, which is a lighter methodology than RL. In addition, by combining visual input with force inputs in the decision-making process, and by adding data augmentation, we can easily handle $16$ different types of pegs and plugs over different locations. Furthermore, we show that our method can handle variability in the environment such as colors or small differences in shape. Our primary contributions are:
\begin{enumerate}[leftmargin=*]
	\item A simple way to formulate an insertion problem in a regression form.
	\item Backward learning, a novel approach to collect data in insertion tasks.
	\item Incorporating data augmentation methods on force and visual inputs.
	\item A rigorous evaluation of the method's performance on (a) 16 different insertion tasks with spatial invariance and (b) real-life assembly tasks with consecutive insertion tasks.
\end{enumerate}



\section{Related work}
\subsection{Contact-rich Manipulation}
Insertion problems have been extensively investigated both by the control community \cite{lozano1984automatic} and the machine-learning community \cite{gullapalli1994learning}.
Insertion may involve a contact model-based control or a contact model-free learning \cite{Xu2019}. Contact model-based strategies build a contact model through analytical tools (static or dynamic), as well as  statistical (GMM, HMM, SVM; \cite{murphy2012machine}) approaches, followed by the delicate design of a corrective strategy, to minimize the contact force. These methods guarantee safety and efficiency and are suitable for special assembly scenarios, but are limited by the ability to model the contact. 

To overcome this limitation, contact model-free strategies have been adopted, where either a Learning from Demonstration (LfD; \cite{schaal1997learning}) or Learning from the Environment (LfE; \cite{kober2013reinforcement}) approach is used. The LfD  approaches benefit from expert demonstrations and enable effective, rapid skill acquisition in physical robots. The recently proposed method for imitating human compliant behaviors using different mapping techniques and without pre-programming is a prime example of LfD's utility \cite{Argall2009}. Similarly to contact model-based approaches, Lfd guarantees safety and efficiency, but its flexibility is limited by the expert demonstrations.

Model-free RL has proved useful for solving challenging tasks in a wide range of problems \cite{sutton2018reinforcement}. However, manipulation tasks are especially challenging due to their continuous action space and their high-dimension state space \cite{kober2013reinforcement}, resulting in large amounts of data collection and long training times \cite{kalashnikov2018qt}. Although recent progress in RL has led to significantly better sample efficiency, algorithms without additional prior knowledge tend to have safety and robot wear-and-tear issues. 

In order to overcome these issues, several approaches tried to exploit prior knowledge on the tasks. For example, residual policy methods use the knowledge on the task to pre-program a fixed policy and used RL only to learn a correction policy. This method was compared to pure RL on a connector insertion task showing improved performance and reduction in learning time\cite{schoettler2019deep}. Another alternative is using LfD to derive a baseline policy \cite{davchev2020residual} or to initiate the learning process \cite{schoettler2019deep,inoue2017deep,luck2019improved}, thereby avoiding some of the safety risk of initial random policy.
The authors of \cite{davchev2020residual}, which used LfD to learn a baseline policy and RL to learn a residual policy correction, showed interesting results in peg, gear and LAN cable insertion. While the above-noted approaches have shown significant results in very challenging problems, online RL algorithms in robotics tend to be impractical in contact-rich tasks.


An alternative approach for enabling deep-RL on robots is Sim2Real: learning in a simulation and transferring the learned policy to a real robot. Such an approach has shown promising results, most notably, in grasping \cite{james2019sim} and, most recently, insertion tasks \cite{spector2020deep}. Sim2Real highlights the generalization ability of RL approaches in insertion tasks if a large amount of diverse data can be collected. However, this approach is limited by the domain expert's ability to model the real-world scenario in the simulation.

\subsection{Multimodal Representation Learning}
Force and visual information are two complementary modalities. While force feedback is statically more informative as the overlap between the peg and hole grows \cite{spector2020deep}, the visual information is diluted as the occlusion increases. On the one hand, high-precision assembly tasks \cite{inoue2017deep} must use force information for the fine-tuning of the manipulation insertion. On the other hand, force information can be useless when there is no overlap between the peg and the hole. This is common in electric plugging (e.g., AC or Aux), where the socket holes are smaller than the system error in localization, control, or grasping. Recent works on insertion problems demonstrated the multi-modal approach's ability to solve flexible and rigid insertion tasks \cite{vecerik2019practical} with a certain degree of generalization \cite{lee2019making}. However, both \cite{vecerik2019practical} and \cite{lee2019making} used fixed cameras to add visual information to the system, thereby increasing the space complexity and decreasing the algorithm's ability to generalize over space. In addition, fixed cameras may insert noise to the system in the form of occlusions. A more suitable approach for insertion is to use a wrist camera at 45 degrees (relative to the End Effector) directed at a point between the two fingers.

\section{Method}
\label{sec:method}
The proposed approach is composed of two steps. The first step entails localizing the holes, sockets, threads, etc. (herein "objectives") in the scene and using a PD controller to follow the calculated path. We refer to this step as "Localization Process". In the second step, a residual policy, denoted by $\pi_{\textrm{residual}}$, is incorporated on top of the main policy. In contrast to other residual policy methodologies, where the policy is added at any state (e.g., \cite{schoettler2019deep},\cite{silver2018residual}), which results in the learning of large actions far away from the hole opening \cite{lee2020guided}, our residual policy is activated only when there is a large gap between the planned kinematics and the actual robot kinematics. In an insertion task, the kinematic gap is due to the collision with the hole surroundings creating a high jerk (or force change). Unlike other methodologies that use only force information in the residual policy (e.g., \cite{spector2020deep}, \cite{davchev2020residual}) or only vision information (e.g., \cite{schoettler2019deep}), we use a multi-modal approach, fusing together force and vision information. In this work, we do not use RL methods to learn the residual policy but rather formulate the problem as a regression problem. This provides us with an offline learning advantage without the risk of damaging the robot or its surroundings. To collect the data efficiently, we use the backward learning approach described in Section \ref{sec:Backward_learning}.

\subsection{The System's Components}
\label{subsec:system_components}
\begin{figure}[b]
\centering
\includegraphics[width=0.25\textwidth]{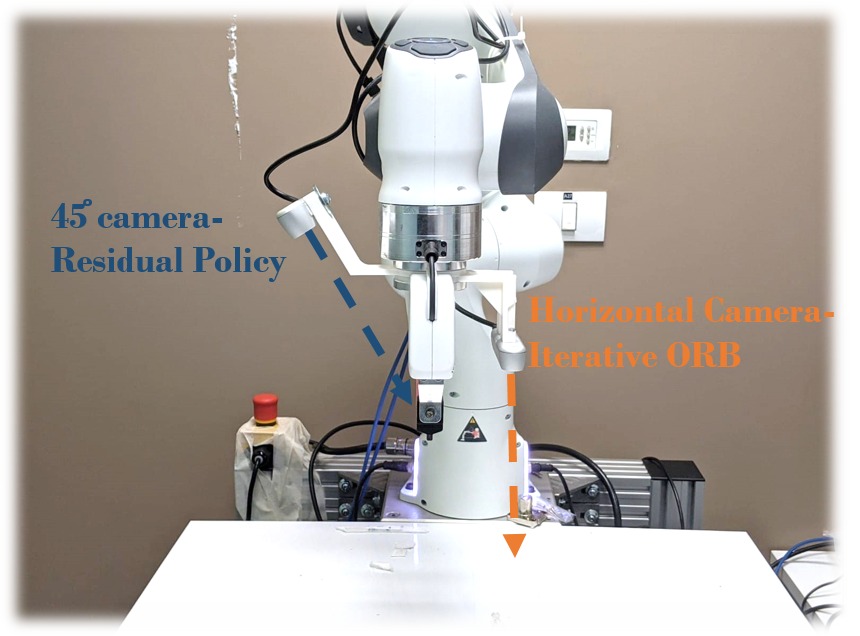}
\caption{\label{fig:Sys} The system's components include 2 RealSense D435 wrist cameras: (1) horizontal w.r.t. to the EEF. (2) tilted in $45^\circ$ and focused at a point between the EFF fingers.}
\end{figure}

We use a Franka Emika Panda arm (Figure \ref{fig:Sys}) with $3$ sensory inputs. The former is a RealSense D435 wrist camera, horizontal w.r.t. to the End Effector (EEF for short); the latter, another D435 camera, tilted in $45^\circ$ and focused at a point between the EFF fingers. The third input is force sensing on the EEF, where we use a force estimator calculated by the robot.

The horizontal camera is needed to localize the objectives, while the $45^\circ$ camera provides fine visual information close to the interaction area. Fused with the force sensing, this combined information stream is conveyed to the residual policy. A scheme of this architecture is presented in Figure \ref{fig:Arch_fusion_yolo}.

\label{sec:system_components}

\subsection{Formulating the Residual Policy in a Regression Form}


The state of the trained agent is composed of two components. The first component is an image of the size $H \times W \times C$, which correspond to the  height, width, and number of channels, respectively. The second component is a vector that describes the force and moment, denoted by $F \triangleq (f_x, f_y, f_z)\in \mathbb{R}^3$ and $M=(m_x, m_y, m_z) \in \mathbb{R}^3$, respectively.

To avoid occlusion and to improve our ability to generalize to different locations, the images are taken from the $45^\circ$ camera (from the robot wrist).
We note that using the camera in such a way offers a crucial advantage, since generalization is easier from the EEF ego perspective than from the fixed cameras. In addition, the input state space is expressed in the EEF coordinate system.

\newcommand{\DX}{\Delta x}
\newcommand{\DY}{\Delta y}
\newcommand{\DZ}{\Delta z}
\newcommand{\DTX}{\Delta \theta_x}
\newcommand{\DTY}{\Delta \theta_y}
\newcommand{\DTZ}{\Delta \theta_z}
\newcommand{\DXd}{\Delta x^d}
\newcommand{\DYd}{\Delta y^d}
\newcommand{\DZd}{\Delta z^d}
\newcommand{\DTXd}{\Delta \theta^d_x}
\newcommand{\DTYd}{\Delta \theta^d_y}
\newcommand{\DTZd}{\Delta \theta^d_z}
\newcommand{\DELTA}{\Delta}

\begin{figure}[!ht]
\centering
\includegraphics[width=0.4\textwidth]{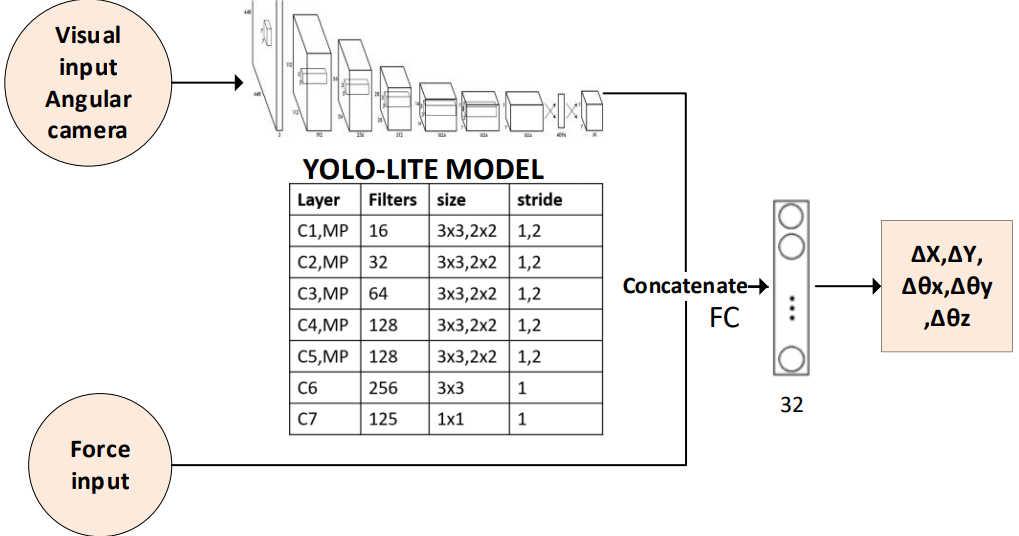}
\caption{\label{fig:Arch_fusion_yolo}Residual Policy Neural Net Architecture}
\end{figure}

The action (i.e., output of the NN) is the real robot's movement in Cartesian space (since it is advantageous in contact-rich tasks \cite{Martin-Martin2019}). Unlike other methods, we express the action as a one-step correction movement that results in success, thereby decoupling the control rate and the action. We denote by $\DX$, $\DY$ and $\DZ$ the desired correction needed for the EEF in the Cartesian space w.r.t. the current location. In addition, $\DTX$, $\DTY$ and $\DTZ$ are the spherical orientation corrections (executed in Quaternion coordinates). We define $\DELTA=(\DX,\DY,\DZ,\DTX,\DTY,\DTZ)$. To summarize, in our regression problem, the algorithm learns the desired $\Delta$ based on the input. 
The objective is to find a policy that maps a tuple of (image, force sensing) $\rightarrow$ action (i.e., $\DELTA$). Formally, we look for $\pi_{\textrm{residual}}: \mathbb{R}^{H\times W \times C} \times \mathbb{R}^{6} \rightarrow \mathbb{R}^6$ with an architecture of the type presented in Figure \ref{fig:Arch_fusion_yolo}. For regression, we need to define the desired output. Specifically, we define with $D\triangleq(\DXd, \DYd, \DZd, \DTXd, \DTYd, \DTZd)$ the labelled (or desired) action to be applied. Therefore, the loss is defined by
\begin{equation}
    loss \triangleq \| \DELTA - D\|^2.
\end{equation}
Below we describe how to collect $D$ (Section \ref{sec:Backward_learning}).

\begin{figure}[b]
\centering
\includegraphics[width=0.48\textwidth]{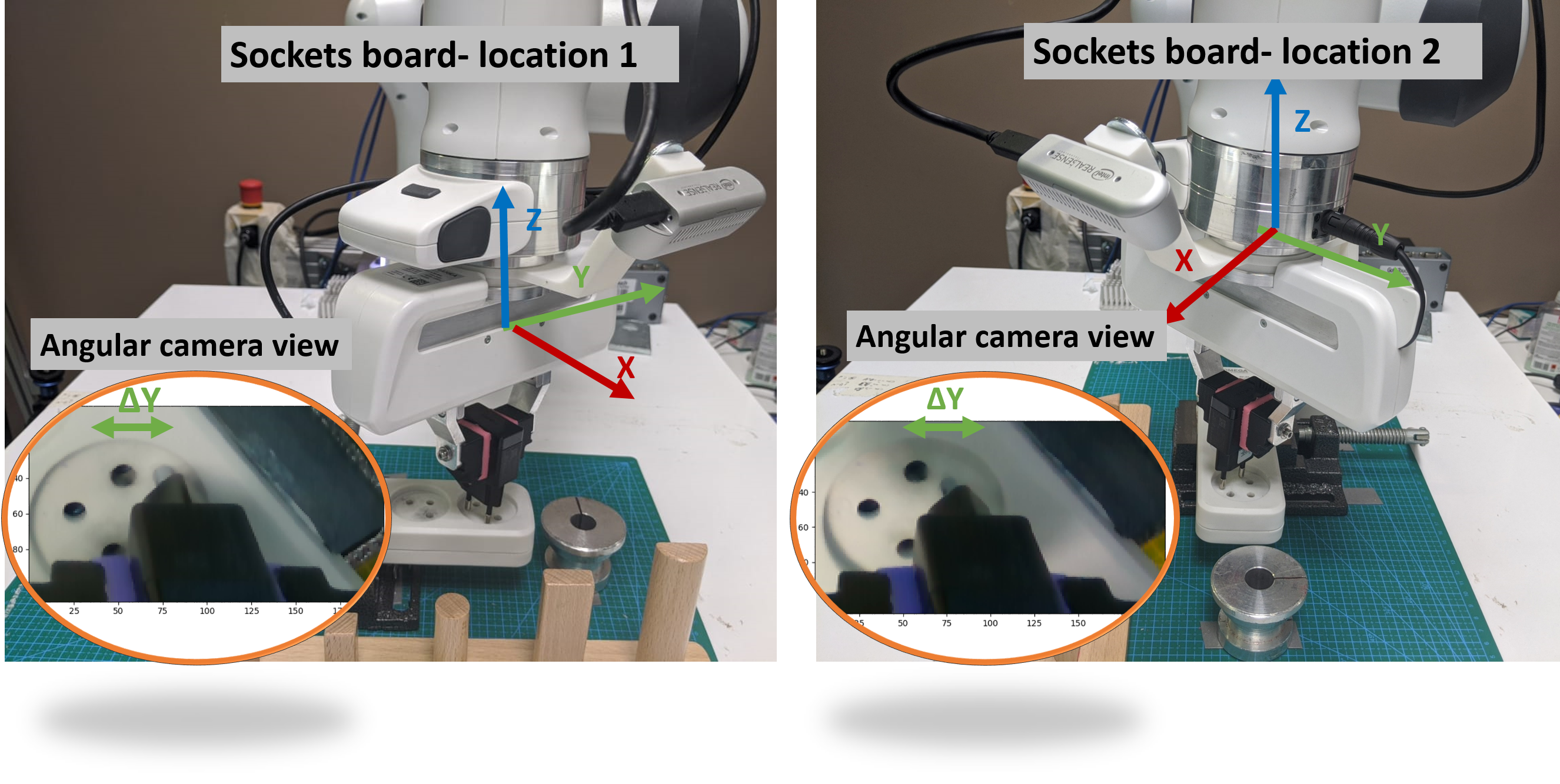}
\caption{Different angles and locations of a socket board with the same $\Delta Y$ in the EEF coordinate system. In the corresponding ellipses, the tilted $45^\circ$ camera is similar. Note that the state is different when the image is captured from a fixed camera, but from the tilted $45^\circ$ camera the captured image is almost identical. In addition, the action when represented in the EEF coordinate system is similar.}
\label{fig:Agnostic}
\end{figure}

In Figure \ref{fig:Agnostic}, we demonstrate the importance of choosing the EEF coordinate system in combination with the $45^\circ$ camera. First, if one opts for a fixed camera, the two different positions presented would result in two different states, with the only difference being in the POV. Second, if one chooses the robot coordinate system instead of the EEF, the action in the two cases would be different.

\subsection{Backward Learning for Data Collection}
\label{sec:Backward_learning}
In order to collect the data for the insertion regression problem, we propose to do a \emph{Backward Learning}, where we initiate our data collection process by putting the robotic arm in the final state; namely, in a state where the peg or plug is positioned inside the hole. We record this position and denote it by $L\triangleq[L_\textrm{pos},L_\textrm{ang}]$, where $L_\textrm{pos}\triangleq(x_0, y_0, z_0)$ and $L_\textrm{ang}\triangleq(\theta^x_0, \theta^y_0, \theta^z_0)$. Now that we have the final state recorded, our goal is to generate collisions around the hole in order to learn $\DELTA$. 

We define $\xvec \triangleq(x,y,z)$ and $\thvec\triangleq(\theta_x, \theta_y, \theta_z)$. Let us define the close proximity to the goal state $L_\textrm{pos}$ as a disk $B_{xy}\triangleq \{ (x,y) : \|(x,y)-(x_0,y_0)\|_\infty < b_0\}$, where typically $b_0=10[\textrm{mm}]$. For $\thvec$, we define the rectangle $B_{\theta}\triangleq \{ \thvec: \|\thvec-L_{\textrm{ang}}\|_\infty < c_0\}$, where typically $c_0=10^\circ$. We note that we chose $b_0=10[\textrm{mm}]$ since a typical visual localization error is no more than $5[\textrm{mm}]$ and no more than $10^\circ$.

Next, we uniformly sample target points from $B_{xy} \times B_\theta$ and denote them with $\{p_i\}_{i=0}^{N_P}$. We define  $F_{\textrm{th}}$ as a force threshold and $M_{\textrm{th}}$ as a moment threshold, both measured on the EEF.
For each $p_i$, first the robot reaches an area above the hole $p_i+z_{max}$ then it tries to reach $p_i$ while measuring the force $F$ and moment $M$ along the trajectory. We note that the arm does not necessarily reach $p_i$, due to collisions with the hole surface. We determine whether a collision occurs during the maneuver (i.e., $F\ge F_{\textrm{th}}$ or $M\ge M_{\textrm{th}}$) and if so, calculate the state and the action. Specifically, for the state, we measure (1) the registered $45^\circ$ camera image, and (2) the force and moment vector. For the action, we calculate a corresponding $\{D_i\}_{i=0}^{N_p}$, where $D_i \triangleq (x^i, y^i, z^i, \theta_x^i,\theta_y^i, \theta_z^i) - L$, i.e., each $D_i$ is a corrective term that needs to be applied in order to bring the EEF to the correct position and pose. To decrease the wear and tear on the robot we choose low values for $F_{\textrm{th}}$ or $M_{\textrm{th}}$ which result in delicate collision with the hole surroundings.
Summarizing, we have input samples $\{p_i\}_{i=0}^{N_P}$ that we wish to regress to output corrective actions $\{D_i\}_{i=0}^{N_p}$. This procedure is presented in Algorithm \ref{Algo:backward}.

The motivation for the proposed Algorithm \ref{Algo:backward} and Procedure \ref{Algo:Generate} is as follows. In insertion tasks, the common errors that result in collision are due to grasping misalignment and errors in visual localization. A grasping misalignment can create a delta error in $\theta_x$ and $\theta_y$ in the EEF coordinate system between the hole's plane and the peg, 
Errors in the localization mainly stem from the visual localization, specifically, in the $x$, $y$, and $\theta_z$. 


{
\SetAlgoNoLine
\begin{algorithm}[!ht]
\caption{Backward Learning for Data Collection}\label{Algo:backward}
    \DontPrintSemicolon
    \SetKwFunction{Fupdate}{update\_tree}
    \SetKwFunction{Ftop}{top\_actions}
    \KwIn{Maximum Height $z_{\textrm{max}}=50[\textrm{mm}]$; Force threshold $F_{\textrm{th}}$; Momentum threshold $M_{\textrm{th}}$}
    \textbf{Init:} Insert peg/plug into the hole, save final pose $L$ \\
    \textbf{Set:} $(\xvec_0, \thvec_{0})=L$, $\textrm{States}=\{\}$, $\textrm{Actions}=\{\}$ \\
    \For{$i \leftarrow 1,\ldots,N_p$}
    {
        $(\xvec^i,\thvec_i) = RDG(\xvec_0,\thvec_{0}) \quad\quad$ \# Procedure \ref{Algo:Generate}\\
        $T=(\xvec_i^x,\xvec_i^y,\xvec_0^z, \thvec_i^x, \thvec_i^y, \thvec_i^z)\quad$ \# $\xvec_0^z$ located in hole\\
        Move robot to state $T+z_{\textrm{max}}$ \# to start point\\
        Plan trajectory and move robot to state $T$ \\ 
        \While{Robot did not reach $T$} {
            Measure force $F$ and moment $M$ \\
            \If{$F \ge F_\textrm{th}$ or $M \ge M_\textrm{th}$} {
                Measure $(x,y)$ and $(\theta^x, \theta^y, \theta^z)$ \\
                $D_i = \{x , y , \theta^x, \theta^y, \theta^z\}$ \\
                $\textrm{States} \leftarrow \{ 45^\circ$ camera$, F, M \}$ \\
                $\textrm{Actions} \leftarrow \{L-D_i\}$ \\
                break while loop \\
           }        
        }
    }
    \Return $\textrm{States}$, $\textrm{Actions}$ \\
\end{algorithm}
\SetAlgorithmName{Procedure}{}{}

\begin{algorithm}[!ht]
\caption{Random Data Generator (RDG)}\label{Algo:Generate}
    \DontPrintSemicolon
    \KwIn{Maximum Randomized Location $b_0=10[\textrm{mm}]$;
    Maximum Randomized Angle $c_0=10^\circ$; 
    Intial pose $\xvec_0$; 
    Intial orientation $\thvec_0$}
        $\xvec \sim \textrm{Uniform}[\xvec_0 \pm b_0]$ \\
        $\thvec \sim \textrm{Uniform}[\thvec_0 \pm c_0]$ \\
        return $\xvec,\thvec$
\end{algorithm}

\SetAlgorithmName{Algorithm}{}{}

\SetAlgoNoLine
\begin{algorithm}[b]
\caption{\strut Residual Policy (RP)}\label{Algo:Residual}
    \DontPrintSemicolon
    \KwIn{Target Point $T$; Force/Momentum thresholds $F_{\textrm{th}}$/$M_{\textrm{th}}$;
    Maximal Time $t_f=10[s]$; Desired force $f_{\textrm{desired}}[\textrm{N}]$; Compliance constant c[m/N]}
    \textbf{Init:} $Policy=\textrm{Base}$ , $NextPoint=T$\\
     \While{Robot did not reach $T$ and time$<t_f$} {
            Robot move to $NextPoint$ \\
            Measure force $F$ and moment $M$ \\
            \If{$F \textrm{ or } M \ge F_\textrm{th},M_\textrm{th} \textrm{ or } Policy=\normalfont{\textrm{Residual}}$} {
                $Policy=\textrm{Residual}$\\
                $CurrentPoint =$ Measure $(x,y,z,\theta^x, \theta^y, \theta^z)$ \\ 
                $\Delta=\pi_{residual}(45^\circ$ camera, $F, M) $ \\
                $\delta_z= -c(f_{\textrm{desired}}-f_z)$\\
                $NextPoint=CurrentPoint+\Delta+\delta_z$ \\
           }        
        }
        
\end{algorithm}
}
\subsection{Data augmentation}
\label{sec:augmentation}
In order for this algorithm to be applicable to an industrial environment, it should have generalization capabilities as well as robustness to small variations in the environment, e.g., colors, shapes, illuminations, etc. To overcome these challenges, we introduce various augmentation processes to the visual input as well as the force input. 

\textbf{Visual Augmentation:} We divide visual augmentations into color and shape augmentations. To overcome variation in peg color or variance in the illumination, we combined three different visual augmentation algorithms. The first is \emph{random convolution}, which was first introduced in \cite{lee2019network}, augments the image color by passing the input observation through a random convolutional layer. The second is \emph{color jitter}, which converts the RGB image to a HSV colormap and adds noise to the HSV channels, resulting in perturbations of brightness, contrast, and color. The third augmentation algorithm, \emph{gray scale}, converts RGB images to grayscale. 

In order to overcome variation in the peg shape, we propose two shape augmentations. The first is \emph{translation}, where a random translation renders the full image within a larger frame, resulting in the original image appearing in a random position across the larger frame. For example, a 100×100 pixel image could be randomly translated within a 108×108 empty frame. The second is \emph{cropping}, in which a random rectangle extracts a random patch from the original frame.

\begin{figure}[b]
\centering
    \includegraphics[width=0.32\textwidth]{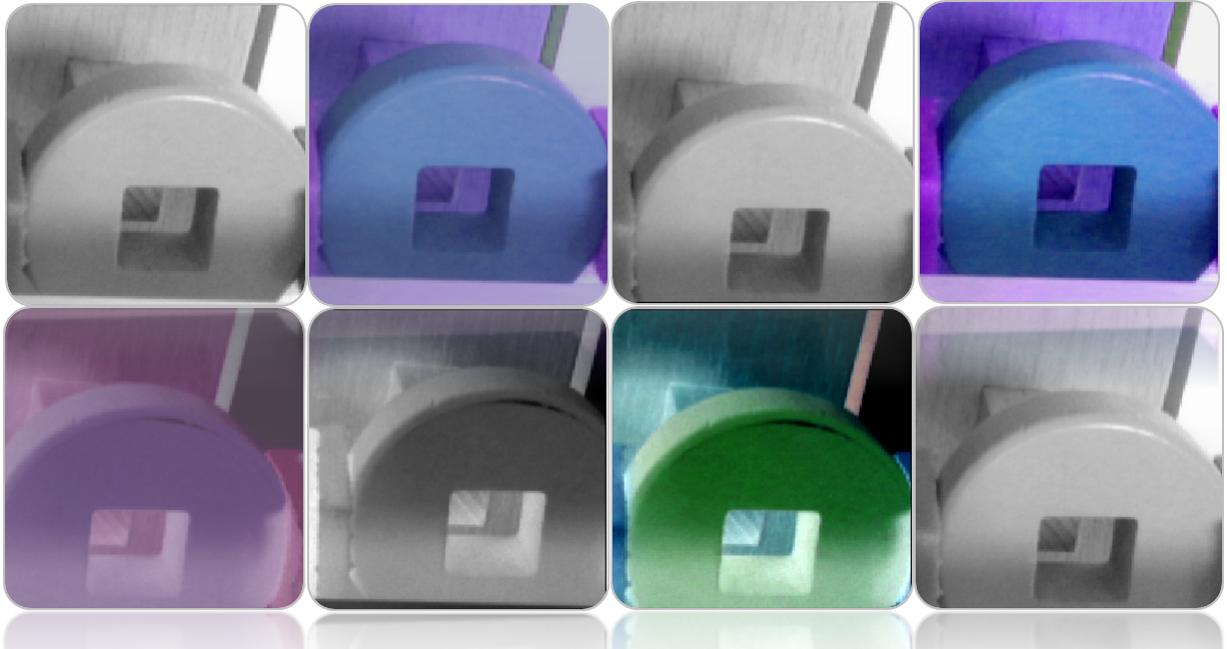}
    \caption{\label{fig:visual_aug} Five different visual augmentations used  simultaneously to tackle color, robustness to illuminations and  variance in shapes. Out of 8 images 4 are in gray scale 4 used color jitter. All images are with different cropping and translation.}%
\end{figure}

All these augmentations are used simultaneously, as presented in Figure \ref{fig:visual_aug}.
We note that since the peg's shape influences the wrist camera's position and distance from the holes in the collision, cropping and translation is required to compensate for these errors.

\textbf{Force and Moment Augmentation:}
The force and moment measurements are strongly dependent on the collision force, which is a product of the main policy velocity, and the transition constant. In addition, when analyzing a simple scheme, one arrives at the conclusion that the direction of vectors(F,M), not their magnitude, is the most valuable factor in insertion tasks. Therefore, we incorporate force augmentations, achieved by simply multiplying both the force and the moment by a random constant $\alpha$: 
$F_{\textrm{augment}}=\alpha F,\quad M_{\textrm{augment}}= \alpha M \quad \alpha \sim U[0,1].$. To the best of our knowledge, we are the first to do so in such a setup.


\subsection{Solving the Angular Camera Ambiguity Problem}
Recovering the distance between two points in world coordinates without depth information using a single image is an ill-posed problem that results in ambiguity \cite{eigen2014depth}. Recently, convolutional networks demonstrated depth prediction from a single image by learning priors about objects and their shapes \cite{eigen2015predicting}, \cite{liu2015learning}. However, single image methods are limited in their ability to generalize to previously unseen types of images. 
This ambiguity can be solved if the depth from the camera is known. But realizing depth sensing in insertion tasks is rather challenging, because most sensors lack the resolution needed to distinguish between two objects from a relatively close distance. Moreover, even if such sensors can be found, they do not resolve the occlusions issues during the task.
We propose to capture each input image only when the plug touches the hole surface. In this scenario, the hole opening and the tip of the peg are aligned on a single plane and the network can easily learn the distance using only a single RGB image.

\subsection{Rough Localization Using Iterative ORB Algorithm}
\label{sub_sec:orb}
Localization can be a computationally intensive task. However, industrial use cases are often limited in their variability and predictability. Moreover, in assembly tasks, for example, products are often composed of a few components connected to a rigid board. We use the ORB algorithm \cite{rublee2011orb} due to its simplicity and agility. In our extensive experiments, we found that in most cases there is no need for any more-advanced localization algorithms.


This rigidity assumption reduces the complexity of the localization problem to the task of finding the board's center, location and orientation. Afterward, the locations of the sockets and electric components relative to the center are easily retrieved. 
We used the ORB algorithm to compare the current image with an image taken a-priory. Before taking this image, we localized each important component in the environment, calculating the delta between the components and the EEF's location (when the pre-taken image was captured). To avoid camera calibration, this process is applied iteratively in a closed loop until the two images match. When the two images match, the delta between the EEF location and the holes should match (with some localization error) to the pre-calculated delta. Thus, we can retrieve the holes location by summing the current EEF position and the pre-calculated delta.


\section{Experiments}
In this section, we describe the experiments conducted to validate our methodology, in which we address the following questions:
\begin{enumerate}
    \item Scalability - How does our algorithm scale to various insertion problems? How many samples are needed? What is the average time point when the object should be inserted?
    \item Spatial Invariance - Can we obtain similar results in other locations?
    \item Generalization - Can we generalize to similar types of plugs with different colors and shapes?
    \item Transferability - How well does the trained policy transfer to other tasks?
\end{enumerate}


\subsection{Experimental setup}

\textbf{Data Collection and Training:} In the first stage, we collect 100 data points (5 minutes for each plug) using the backward learning approach (Algorithm \ref{Algo:backward}). In the second stage, we train a residual policy neural net architecture for each plug (described in Figure \ref{fig:Arch_fusion_yolo})  with the proposed visual and force augmentations of Section \ref{sec:augmentation}. Based on 100 samples, we generate 640,000 training samples (10,000 repeats on 64 samples batches) using 2 GPU GeForce RTX 2080 Ti. The total training time takes $40$ minutes.

\textbf{Testing Procedure:} We test our method in the system described in Section \ref{subsec:system_components}, following a similar procedure to the one described for Algorithm \ref{Algo:backward}. In each experiment, we examine one plug-socket setup. We generate test points, denoted by $N_{\textrm{test}}$, around the testing socket in a similar manner to lines 5-9 in Algorithm \ref{Algo:backward}, by uniformly sampling $|N_{\textrm{test}}|$  target point $T$ (line 9) from $B_{xy} \times B_\theta$.  We note that $T$ is the localization point and that it may contain errors from grasping, misalignment and localization. Afterward, the robot sends $T$ to the process described in Algorithm \ref{Algo:Residual}. 
The PD controller executes each movement command for 0.4[s].
A trial is considered successful if the robot inserts the peg/plug into its designated hole with tolerance $z-z_0<1[\textrm{mm}]$. An unsuccessful trial is when the duration it takes to complete the task exceeds $10[\textrm{sec}]$. We repeat each insertion trial $200$ times (with random initialization) in order to achieve statistically significant results. In order to examine realistic scenarios, the pegs are not fixed to the gripper (allowing slippage between the peg and the fingers) during insertion, and the 200 trials are consecutively applied in a row to intentionally create misalignments during the insertions. 

\begin{figure*}[ht]
\centering 
\includegraphics[width=\textwidth]{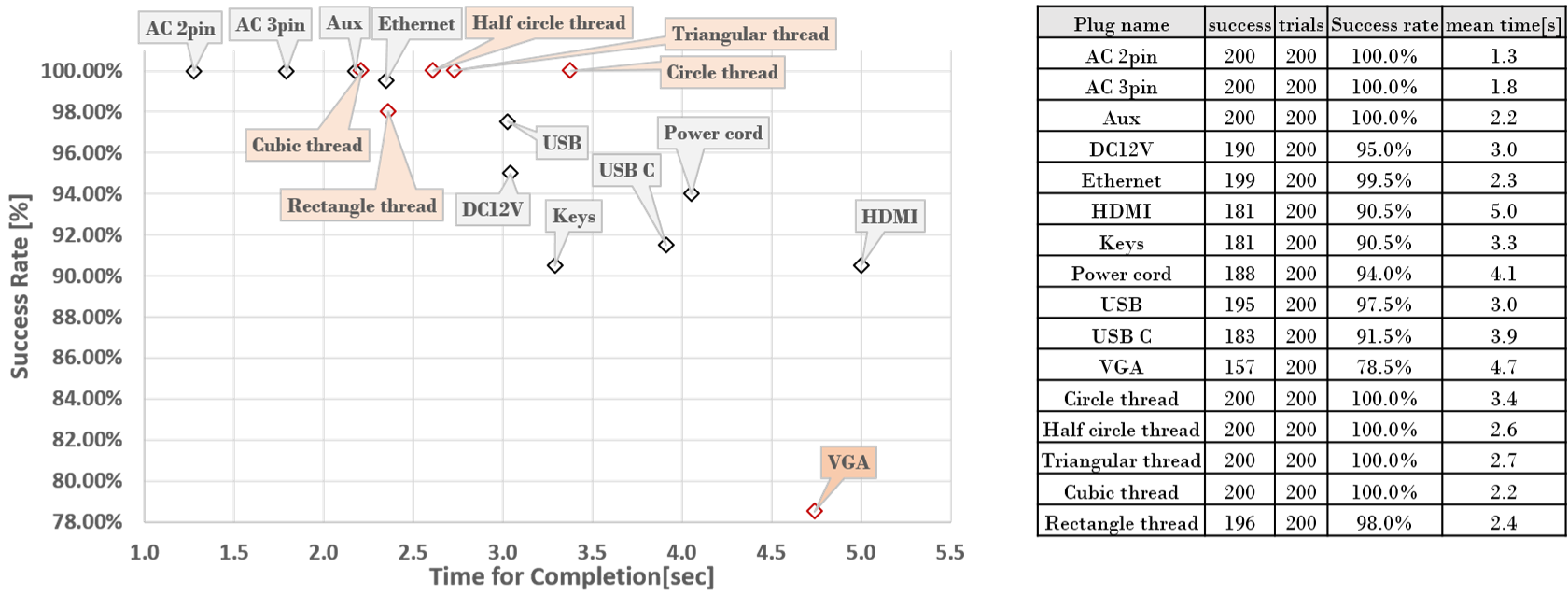}
\caption{Overall performance on the 16 insertion and threading tasks averaged over 200 trials, in a graph (left figure) and in a table (right figure) . In the graph the gray boxes correspond to insertion tasks and the orange boxes correspond to threading tasks.
\label{fig:all_results}}
\end{figure*}

\subsection{Scalability for different insertion setups}
\label{sec:scalability}
In order to evaluate our method's ability to scale to $16$ different tasks, we collected 16 different real-life insertion and threading tasks with different properties. The tasks were split into two groups: insertion tasks and threading tasks. In insertion tasks, the robot holds the peg and tries to insert it into a designated hole. In threading tasks, the robot holds a shaped hoop in its gripper and threads it over a shaft. The insertion tasks include 10 different pegs/plugs (AC 2-pin, AC 3-pin, Aux, DC12V, Ethernet RJ45, HDMI, Keys, Power cord, USB, and USB-C) with different properties that effect the insertion process; for example, different shapes (e.g., non-convex shapes, like an electric plug, vs. convex shapes, like a USB-c plug), clearance sizes (e.g., 0.1mm in HDMI vs. 1mm in AUX) and sizes (2.5mm in Aux vs. 20mm in power chord). The threading tasks include $6$ different shapes: $5$ wooden shafts (triangular, cubical, cylindrical, rectangular, and half circle) and one plug (VGA). In addition, the threading problems varied in many properties, such as friction (wood vs. steel) shape (triangular, cubical, etc.) or clearness (0.1mm for the cubical shapes, 1mm for the triangular, etc.).

Figure \ref{fig:all_results} summarizes the success rate and time for completion of the 16 different insertion cases, averaged over 200 trials. The insertion-oriented tasks (marked gray) scored 95.9$\%$, on average, over all the tasks. The lowest score in the insertion tasks was 90.5$\%$, in the keys and HDMI tasks, while the highest score was $100\%$, in the AC 2-pin, AC 3-pin and Aux. The insertion-oriented tasks' average trial duration was 2.9[s]. The longest time duration was $5[\textrm{sec}]$, with the HDMI plug, and the shortest was $1.3[\textrm{sec}]$, with the AC2 pin. The threading-oriented tasks (marked orange) scored an average of $96\%$ over all the tasks. Here, the lowest score was $78.5\%$, in the VGA task, and the highest was 100$\%$, in the circle, half circle, triangle, and cubical wooden threads. The threading-oriented tasks' average duration was $3.0[\textrm{sec}$ over all the tasks. The longest duration was $4.7[\textrm{sec}$, with the VGA plug, and the shortest was $2.2[\textrm{sec}$, with the cubical thread.

During the experiments, we noticed three reasons for the variance in the success rate and time. First, the methods encountered difficulties in small clearance, necessitating force (such as HDMI and a power cord). In those cases, small errors might result in jamming. Second, another challenging type of scenario was when the hole and clearance were small, such as in USB-C and DC12V. Those cases require very fine control, as the robot can move over the hole without inserting the peg. Third, the method had a difficulty tackling an insertion where the hole and the plug are of similar color, such as a HDMI and power cord. In these cases, visually separating between the plug and socket is hard. 
The VGA task was, by far, the task with the lowest scores. During the experiments, we noticed that the combination of small clearance in the VGA with the high friction between the peg and the hole surface results in many instances of the peg becoming jammed or the controller getting stuck due to friction.  

\subsection{Sample efficiency}
\begin{figure}[ht]
\centering
\includegraphics[width=0.4\textwidth]{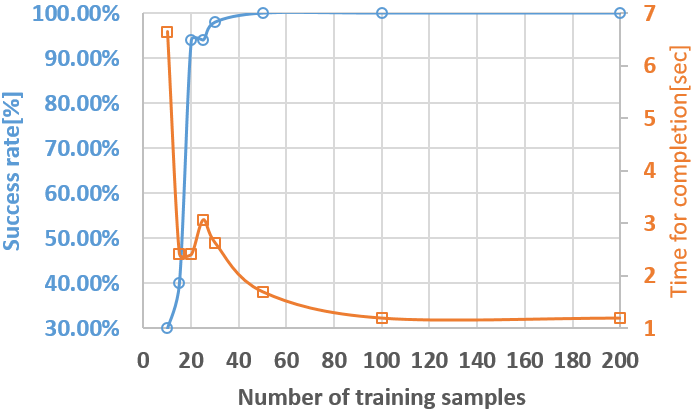}
\caption{\label{fig:number_of_samples} The success rate (blue, circles) and time for completion (orange, squares) given number of training samples. Given enough training samples we see that the system achieves the minimal time for plugging and $100\%$ success rate.}
\end{figure}

In this section we assess how many samples are required to successfully accomplish an insertion task. In order to calculate the relation between the number of samples and the success rate or time for completion, we trained $8$ different policies using $10$, $15$, $20$, $25$, $30$, $50$, $100$ and $200$ samples. Each policy was tested $50$ times on the AC 2-pin plug task. The success rate and time for completion vs. samples required is depicted in Figure \ref{fig:number_of_samples}. The  success rate rapidly increases between $0$ and $50$ samples, reaching 100$\%$ in 50 data points, while The time for completion steeply declines between $0$ and $100$ samples, reaching a minimum of $1.3[\textrm{sec}]$ at $100$ samples. Based on this evidence, we can re-verify that $100$ samples are sufficient for the  Section \ref{sec:scalability} experiments.

\subsection{Generalization}
\begin{figure}[ht]
\centering
\includegraphics[width=0.4\textwidth]{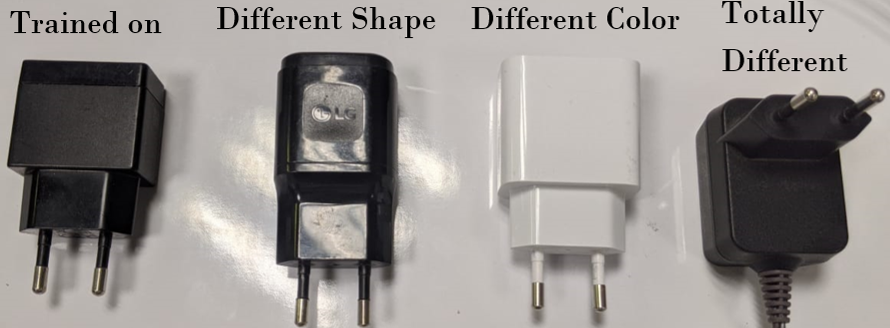}
\caption{\label{fig:diffrent_shape_color} Closely related problems. Four different electric plugs demonstrating real-life variations in shape, color, and other properties.}
\end{figure}

\textbf{Generalize Over Location and Orientation:}
To evaluate our method's ability to generalize over space, we preformed the AC 2-pin plugging in four different locations and orientations, by placing the socket board on the four corners of the robot table. Each experiment was repeated 50 times and the success rate and time for completion were measured. The results indicate no decline in the success rate, but the experiment took 30$\%$ longer, on average, than in the learned location experiment.\\
\textbf{Generalize Over Shape:}
To evaluate our method's ability to generalize over plug shapes, we trained a policy with an AC 2-pin plug and tested the performance on different AC 2-pin plug shapes( figure \ref{fig:diffrent_shape_color}). Our results indicate a minor decline ($1$ unsuccessful experiment) in the success rate and an $60\%$ increase, on average, in the duration time in comparison to the basic plug. We note that without visual augmentation our model success rate drops to $41\%$.  \\
\textbf{Generalize Over Color:}
To evaluate our method's generalization over colors, we trained a policy with a black AC 2-pin plug and tested the performance on a white AC 2-pin plug( figure \ref{fig:diffrent_shape_color}). The results indicate a minor decline (two unsuccessful experiments) in the success rate and a 100$\%$ increase, on average, in the duration time where without visual augmentation our model success rate drops to $18\%$. \\
\textbf{Transferability:} 
In this section, we evaluated how well a trained policy is transferred to other tasks. The policy was trained over $100$ samples on the AC 2-pin task. We evaluated how many additional samples from the new task are needed to accomplish the task. The new task was chosen so as to be similar to the trained task in order to be able to examine whether the algorithm captures the commonality between the tasks. We chose as the new task a 2-pin plug as well, but it required a different grasping method and had a significantly different shape( figure \ref{fig:diffrent_shape_color} right plug). We trained $5$ different policies: adding  $0$, $5$ ,$10$, $15$ and $20$ samples. Trivially, the algorithm without any additional sample, as expected, score $0\%$. Notably, only five samples were required to achieve a score of $70\%$, and $10$ for a score of $100\%$, in comparison to $50$ without the pre-training policy. This implies that roughly only 30[sec] of data collection is needed to learn new insertion tasks using our method.

\section{End-to-end assembly tasks}
\label{sec:end2end}
In this section, we tested our method on complex assembly tasks, from grasping to hole localization, to insertion. The tasks goal was to test the method ability to learn a variety of insertion tasks with only one policy, to handle consecutive insertion tasks even in close proximity, and to combine both the localization process and the insertion process. In the following, we successfully demonstrate these abilities on two environment designs.

\noindent \textbf{First Environment Design} (Figure \ref{fig:scalable_solution}b up): A board with $3$ AC sockets, $2$ USB sockets, and an Ethernet socket is presented. The robot's goal is to insert all the plugs into their designated holes, one after the other, with only one policy. We emphasize that after the insertion of each plug, the plug is left in the hole, which changes the environment's state and disturbs the insertion of the other plugs. 
In order to evaluate our method's robustness and generalization, the plugs were of different colors and shapes; in addition, the USB and Ethernet cord remained attached, thus creating occlusions.

\noindent \textbf{Second Environment Design} (Figure \ref{fig:scalable_solution}b bottom): A wooden threading game that includes five different wooden shafts and 20 rings with matching holes. The goal of the game is to insert the different shapes into their matched shafts. This game included the following pieces: 5 half circles, 4 rectangles, 3 circles, 2 squares and a triangle. To solve the environment, the robot needs to insert all 15 pieces into their matched holes.

\noindent \textbf{Training Procedure:} We begin with collecting 100 data points from each assembly component (e.g., 300 data points for the first environment) in order to learn a general residual policy. Afterward, we save the holes' and shafts' locations in advance and register an image (together with the EEF location) using the horizontal camera for the localization process (section \ref{sub_sec:orb}). Finally, we place the plugs on a conveyor, recording each of the plugs' location for grasping purposes.

\noindent \textbf{Testing Procedure and results:} 
Before starting the process, we randomly place the socket board on the robot table. Then, the localization process (as describe in Section \ref{sub_sec:orb}) starts by roughly localizing the holes and saving each of their locations. After localization, the robot enters an operational loop, in which it reaches, grasps, and inserts a plug into its designated hole, until all the objects are inserted. In both settings, all plug-ins were successfully inserted.

\section{Conclusions}
\label{sec:conclusions}
We introduced a novel framework for learning a wide range of insertion tasks. We address the wear-and-tear and safety issues in contact-rich tasks by formulating the residual policy in a regression form. In addition, we introduce a backward approach for the fast collection of data without endangering the robot. In addition, we used a multimodal approach that combines vision and force sensing but replaces the commonly used fixed camera with a 45 deg wrist camera for better generalization.

Our results show that an agent can reliably solve 16 different real-life insertion tasks and is robust to socket pose and variance in plug color or shape. Finally we demonstrate our framework's ability to tackle two real-life assembly tasks with multi-insertion objectives. Qualitative results can be found in our supplementary video on our website: \href{https://sites.google.com/view/insertionnet/}{https://sites.google.com/view/insertionnet/}

\section*{ACKNOWLEDGMENT}

We thank the anonymous reviewers and Mrs. Dana Rip for their very helpful comments that improved this manuscript.

\bibliographystyle{IEEEtran}
\bibliography{IROS}
\newpage

\vfill




\end{document}